%% file: root.tex
\documentclass[letterpaper, 10 pt, conference]{ieeeconf}  % Comment this line out if you need a4paper

\IEEEoverridecommandlockouts                              % This command is only needed if 
\usepackage{epsfig} % for postscript graphics files
\usepackage{mathptmx} % assumes new font selection scheme installed
\usepackage{times} % assumes new font selection scheme installed
\usepackage{amsmath} % assumes amsmath package installed
\usepackage{amssymb}  % assumes amsmath package installed
\usepackage{graphicx}
\usepackage{layouts}
\usepackage{booktabs}
\usepackage{sidecap}
\usepackage{caption}
\usepackage{siunitx}
\usepackage{multirow}
\usepackage{bbm}
\usepackage{xcolor}
\usepackage{tikz}

\DeclareSIUnit{\nothing}{\relax}

\newcommand{\red}[1]{#1}

\title{\LARGE \bf
Continual \red{Adaptation} of Semantic Segmentation\\using Complementary 2D-3D Data Representations
}
\author{Jonas Frey$^{1}$, Hermann Blum$^{2}$, Francesco Milano$^{2}$, Roland Siegwart$^{2}$, Cesar Cadena$^{2}$ % <-this % stops a space
\thanks{$^{1}$ Robotic Systems Lab, ETH Zurich, Switzerland; $^{2}$ Autonomous Systems Lab, ETH Zurich, Switzerland; This research is supported by the HILTI group and the European Union’s Horizon 2020 research and innovation programme under grant agreement No 101017008.}
}
\newcommand{\ra}[1]{\renewcommand{\arraystretch}{#1}}

\begin{document}

\maketitle
\thispagestyle{empty}
\pagestyle{empty}

\input{content/0_Abstract.tex}
\input{content/1_Introduction.tex}

\input{content/2_Related_Work.tex}
\input{content/3_Approach.tex}
\input{content/4_Implementation_Details.tex}
\input{content/5_Experiments.tex}
\input{content/6_Conclusion.tex}

\bibliography{ references}

\end{document}

%% file: content/0_Abstract.tex
\begin{abstract}
Semantic segmentation networks are usually pre-trained once and not updated during deployment. As a consequence, misclassifications commonly occur if the distribution of the training data deviates from the one encountered during the robot's operation. We propose to mitigate this problem by adapting the neural network to the robot's environment during deployment, without any need for external supervision. Leveraging complementary data representations, we generate a supervision signal, by probabilistically accumulating consecutive 2D semantic predictions in a volumetric 3D map. We then train the network on renderings of the accumulated semantic map, effectively resolving ambiguities and enforcing multi-view consistency through the 3D representation. In contrast to scene adaptation methods, we aim to retain the previously-learned knowledge, and therefore employ a continual learning experience replay strategy to adapt the network. Through extensive experimental evaluation, we show successful adaptation to real-world indoor scenes both on the ScanNet dataset and on in-house data recorded with an RGB-D sensor. Our method increases the segmentation \red{accuracy} on average by \red{9.9\%} compared to the fixed pre-trained neural network, while retaining knowledge from the pre-training dataset.
\end{abstract}

%% file: content/1_Introduction.tex
\section{INTRODUCTION}
Robotic perception tasks such as semantic segmentation or object detection rely on large-scale neural networks, which require gathering and annotating large datasets to be trained.
This tedious process is costly, time-intensive, and error-prone.
Furthermore, a dataset captured at a fixed point in time cannot cover every possible data point in the future for complex tasks in unknown environments.
This leads to the common problem of a distribution mismatch between the available labeled training data (source domain) and the actual data of interest (target domain) encountered in a robot's mission. 
In semantic segmentation, such domain gaps commonly cause misclassification of small semantic details in favor of the dominant neighboring semantic class~\cite{SemSegReview17} and make it harder to segment a scene correctly from an arbitrary camera viewpoint, in particular under challenging lighting conditions. 
While these limitations call for the need of performing network adaptation in new environments, existing methods either require prior knowledge of the environment~\cite{Blum21} or ground-truth supervision during deployment~\cite{PLOP20}. However, to be performed on autonomous robots, adaptation to a new environment should not rely on external supervision, since this is in general not available in a deployment scenario~\cite{Lesort2020CLForRobotics}. This therefore restricts adaptation methods for robotics to operate in an unsupervised manner.

\input{content/figures/labdata_matrix}

Fortunately, mobile robots can usually observe the same area of their deployment environment from different viewpoints, thus potentially providing the means to resolve semantic ambiguities. This is however not exploited by standard approaches used in robotics for semantic segmentation, which operate on a per-frame basis, producing a prediction for each image separately. 
Small, occluded or partially observed objects may therefore be misclassified due to the lack of sufficient context, but may be correctly classified in consecutive frames with different viewpoints.
While a number of works integrate semantic predictions into a global 3D representation to achieve more consistent labeling of the scene~\cite{voxbloxplusplus19,kimera20,BundleFusion17}, none of these approaches leverages the view consistency to adapt the semantic segmentation network, which in their experiments is only trained once before deployment and not updated during the mission. 
As illustrated in Figure~\ref{fig:labdata_matrix}, this work instead proposes to explicitly leverage multi-view consistency both to increase the robustness of the semantic labels and to adapt the network to a new environment. 
To this extent, we accumulate individual 2D semantic predictions into a 3D map and generate a new training signal for the network  
by reprojecting the fused semantic information back into 2D pseudo-labels.
To the best of our knowledge, we are the first to propose adapting a neural network according to a supervision signal generated by explicitly transforming network predictions between 2D and 3D.
Through extensive experimental evaluation, we show that the multi-view consistency enforced by this supervision signal allows the network to reliably increase its accuracy during deployment. 

Na\"ive adaptation of the network to the generated supervision signal can however cause forgetting of previously-acquired knowledge~\cite{catastrophic_forgetting89}. 
This is undesirable for mobile robots, which have to frequently change operating environments, but can possibly return to previously encountered scenes.
This problem relates to the field of \textit{continual learning}, which studies how previous knowledge can be preserved while new one is integrated into a neural network. 
To achieve the adaptation of the network to the current environment while counteracting forgetting, we adopt an experience replay continual learning strategy~\cite{rethinking_er20}, regularizing the adaptation to the pseudo-labels with stored samples from the pre-training dataset. \red{Consequently, we refer to our continual learning based approach of network adaptation as continual adaptation.}
Contrary to the other methods that explore semantic segmentation in a continual learning setting, our approach is suited for online deployment, exploits a 3D representation to enforce multi-view consistency, and only requires an RGB-D sensor and the associated camera poses. 
We evaluate our method on the indoor, real-world ScanNet~\cite{Scannet17} dataset, \red{showing a remarkable increase of the semantic segmentation accuracy in average by 9.9\%} relative to the static network.  
Additionally, we provide qualitative deployment experiments with a handheld \mbox{RGB-D~sensor}. 

To summarize, our main contributions are the following:
\begin{enumerate}
    \item \red{We propose to render pseudo labels from a semantic map as a self-supervision technique for segmentation;}
    \item We employ the generated pseudo-labels to adapt a semantic segmentation network using continual learning;
    \item We present quantitative evaluation on both a real-world dataset and in real-world experiments using an RGB-D sensor, showing consistent improvement of the prediction accuracy of the adapted network.
\end{enumerate}

%% file: content/figures/labdata_matrix.tex
\begin{figure}[t]
\def\iwidth{0.21\linewidth}
\def\igap{8pt}
\def\rowheight{-1.55}
\centering
\begin{tikzpicture}
\node[rotate=270] at (-1.2, 0) {\scriptsize\strut 1-Pred};
\node[inner sep=0pt] at (0, 0) {\includegraphics[width=\iwidth]{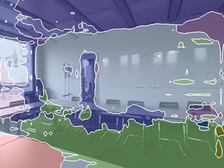}};
\node[inner sep=0pt] at (\iwidth+\igap, 0) {\includegraphics[width=\iwidth]{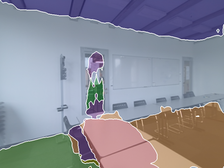}};
\node[inner sep=0pt] at (\iwidth*2+\igap*2, 0) {\includegraphics[width=\iwidth]{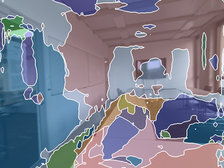}};
\node[inner sep=0pt] at (\iwidth*3+\igap*3, 0) {\includegraphics[width=\iwidth]{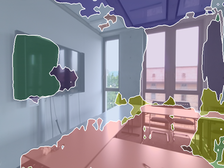}};

\node[rotate=270] at (-1.2, \rowheight) {\scriptsize\strut 1-Pse};
\node[inner sep=0pt] at (0, \rowheight) {\includegraphics[width=\iwidth]{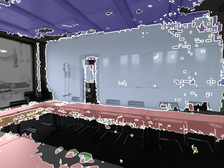}};
\node[inner sep=0pt] at (\iwidth+\igap, \rowheight) {\includegraphics[width=\iwidth]{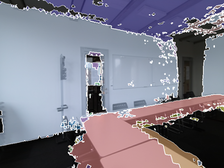}};
\node[inner sep=0pt] at (\iwidth*2+\igap*2, \rowheight) {\includegraphics[width=\iwidth]{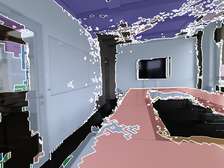}};
\node[inner sep=0pt] at (\iwidth*3+\igap*3, \rowheight) {\includegraphics[width=\iwidth]{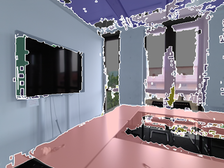}};

\node[rotate=270] at (-1.2, 2*\rowheight) {\scriptsize\strut 2-Pred (ours)};
\node[inner sep=0pt] at (0, 2*\rowheight) {\includegraphics[width=\iwidth]{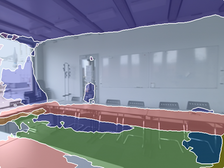}};
\node[inner sep=0pt] at (\iwidth+\igap, 2*\rowheight) {\includegraphics[width=\iwidth]{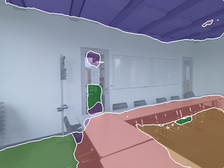}};
\node[inner sep=0pt] at (\iwidth*2+\igap*2, 2*\rowheight) {\includegraphics[width=\iwidth]{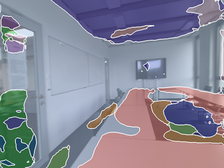}};
\node[inner sep=0pt] at (\iwidth*3+\igap*3, 2*\rowheight) {\includegraphics[width=\iwidth]{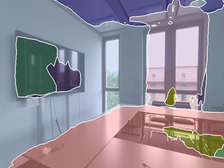}};

\node[rotate=270] at (-1.2, 3*\rowheight) {\scriptsize\strut Ground Truth};
\node[inner sep=0pt] at (0, 3*\rowheight) {\includegraphics[width=\iwidth]{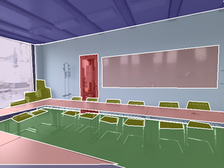}};
\node[inner sep=0pt] at (\iwidth+\igap, 3*\rowheight) {\includegraphics[width=\iwidth]{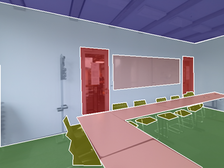}};
\node[inner sep=0pt] at (\iwidth*2+\igap*2, 3*\rowheight) {\includegraphics[width=\iwidth]{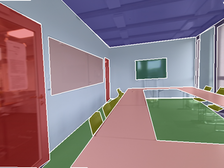}};
\node[inner sep=0pt] at (\iwidth*3+\igap*3, 3*\rowheight) {\includegraphics[width=\iwidth]{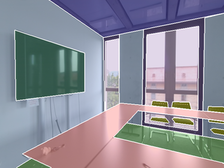}};
\end{tikzpicture}
	\caption{Comparison of semantic segmentation performance on a recording from an indoor scene. From top to bottom: Pre-trained network predictions (1-Pred), rendered pseudo-labels leveraging multi-view consistency (1-Pse), our adapted network predictions after training on 1-Pse with a continual learning strategy (2-Pred), ground truth annotated by us. All semantic classes are color-coded. \red{The continually-learned network produces more view-consistent predictions than the pre-trained network, as visible for instance for the desk.} In the second row black indicates that no semantic class can be determined, due to the missing depth information.}
	\label{fig:labdata_matrix}
	\vspace{-0.6cm}
\end{figure}
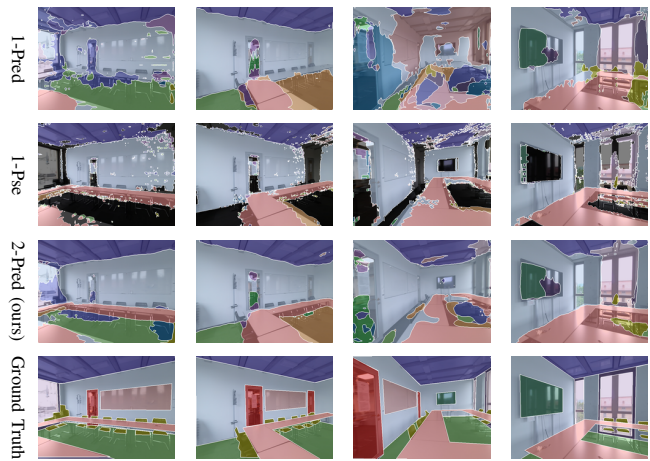

%% file: content/2_Related_Work.tex
\section{RELATED WORK}
\textbf{Fusing semantic information} from different viewpoints to achieve better semantic understanding of a scene is proposed in a number of previous works.
SemanticFusion~\cite{semanticfusion16} leverages a semantic segmentation network and a SLAM system to create a surfel-based representation of the scene,
which probabilistically accumulates semantic information. 
Kimera~\cite{kimera20} uses a similar approach but tracks semantics using a regular voxel grid instead of surfels. 
Voxblox++~\cite{voxbloxplusplus19} identifies individual object instances and organizes them in a volumetric object-centric map. 
More recently,~\cite{3DSceneGraph19} proposed to augment a 3D representation of the environment with hierarchical scene graphs, which encode relationships between elements.
All the above methods, however, rely on a fixed network which is assumed to be accurate and not updated with the semantic information collected in the generated map.
On the contrary, we propose a method that accumulates semantic information similarly to~\cite{voxbloxplusplus19} and~\cite{kimera20}, but continually trains the neural network using the accumulated information about the current environment in an unsupervised manner. 

\textbf{Representation learning} methods aim to create meaningful embeddings of input data points, that can be used for further downstream tasks. 
The majority of the methods in this field leverage unlabeled data, achieving state-of-the-art performance in generating visual representations~\cite{visual_representation_learning19}.
Most methods are studied as an offline pre-training mechanisms~\cite{pretraining19}. 
Here, particularly relevant is the work by~\cite{Pri3D21}, which highlights the benefit of leveraging the view consistency enforced in 3D as a prior for 2D tasks involving semantics. 
However, while this work focuses on offline pre-training of network features, we explicitly integrate semantic predictions in 3D and tackle an online setting with adaptation of the network to new data. 

\textbf{Domain adaptation} methods allow to transfer previously learned information from a source domain to a desired target domain. 
In the context of semantic segmentation the domain gap is often studied from training on simulated data and adapting to real-world data~\cite{Hoffman_cycada2017}. 
Methods can be categorized based on their mode of operation. 
Input-level approaches modify the input to the neural network to mimic the source domain, also referred to as input style transfer~\cite{ROAD17, CrDoCo}. 
Feature-level approaches align the latent feature representation within the neural network between samples from the source and target domain. 
This can be achieved by using a domain classifier network~\cite{ganindomainadversarial16} or adversarial domain adaptation~\cite{Hoffman_cycada2017}, resulting in a high degree of similarity between the features obtained from the target domain and those from the source domain.
At the output level, label statistics are used to adapt the prediction of the network. 
\red{URMA~\cite{2021UncertaintyReduction} performs unsupervised domain adaptation (UDA) for semantic segmentation by reducing the uncertainty in the target domain through auxiliary decoders, which predict multiple segmentation estimates from dropout-corrupted latent feature maps. The multiple predictions are used to compute an uncertainty loss, which is minimized together with the cross-entropy loss of the network predictions to achieve adaptation to the new domain.} 
In general, all domain adaptation methods focus on adapting a network to a specific domain, while discarding the performance achieved on the source domain. In our robotic mission scenario, instead, domains (i.e., scenes) may be revisited and knowledge should thus be retained, rendering most domain adaptation strategies unsuitable for this task. 

\textbf{Continual learning}~\cite{Lesort2020CLForRobotics} focuses on the problem of integrating new knowledge into a network while preventing the accuracy on previously seen data from decreasing significantly when training on a new data distribution, a commonly-observed phenomenon referred to as catastrophic forgetting~\cite{catastrophic_forgetting89}. 
Among the different strategies proposed to address this issue, experience replay has proven to be particularly effective~\cite{er18,rethinking_er20,a-gem18,OGD19}, often outperforming approaches based on more complex designs~\cite{Prabhu2020Gdumb, Knoblauch2020OptimalCL}. 
In experience replay, a subset of previously learned samples is stored and used to regularize the training procedure.
Despite its relevance for real-world settings, where non-stationary data distributions are common, little research is conducted to address robotic settings within the continual learning field. 
The sparse existing literature evaluates on a small scale or unrealistic problem sets~\cite{a-gem18, OGD19, P-MNIST}, and mainly focuses on classification tasks~\cite{iCaRL_cifar100,CORe50,P-MNIST}. 
A limited number of works has explored semantic segmentation in continual learning~\cite{CL_Semseg21,CL_Semseg20,CL_background20,CL_SemSeg_background_shift21,PLOP20}. 
However, these approaches are not designed for application in an online scenario, tackle segmentation purely in 2D on a per-frame basis, and rely on the availability of ground-truth supervision. 
On the contrary, we focus on the setting of online deployment where no ground-truth labels are available, and exploit multi-view consistency across frames, both to increase the robustness of the segmentation and to generate a learning signal for adaptation.
Recently, the authors of \cite{Blum21} proposed to adopt a continual learning strategy to improve the segmentation and localization capability of a construction robot in a self-supervised manner. However, their approach relies on known building meshes and a LiDAR sensor to generate pseudo-labels. Additionally, it is only evaluated on the task of binary classification into fore- and background. Our method is  more versatile, as it performs multi-class semantic segmentation and does not rely on any external knowledge such as precise CAD maps or expensive sensors.

%% file: content/3_Approach.tex
\section{APPROACH}

\begin{figure}[t]
	\includegraphics[width=\linewidth]{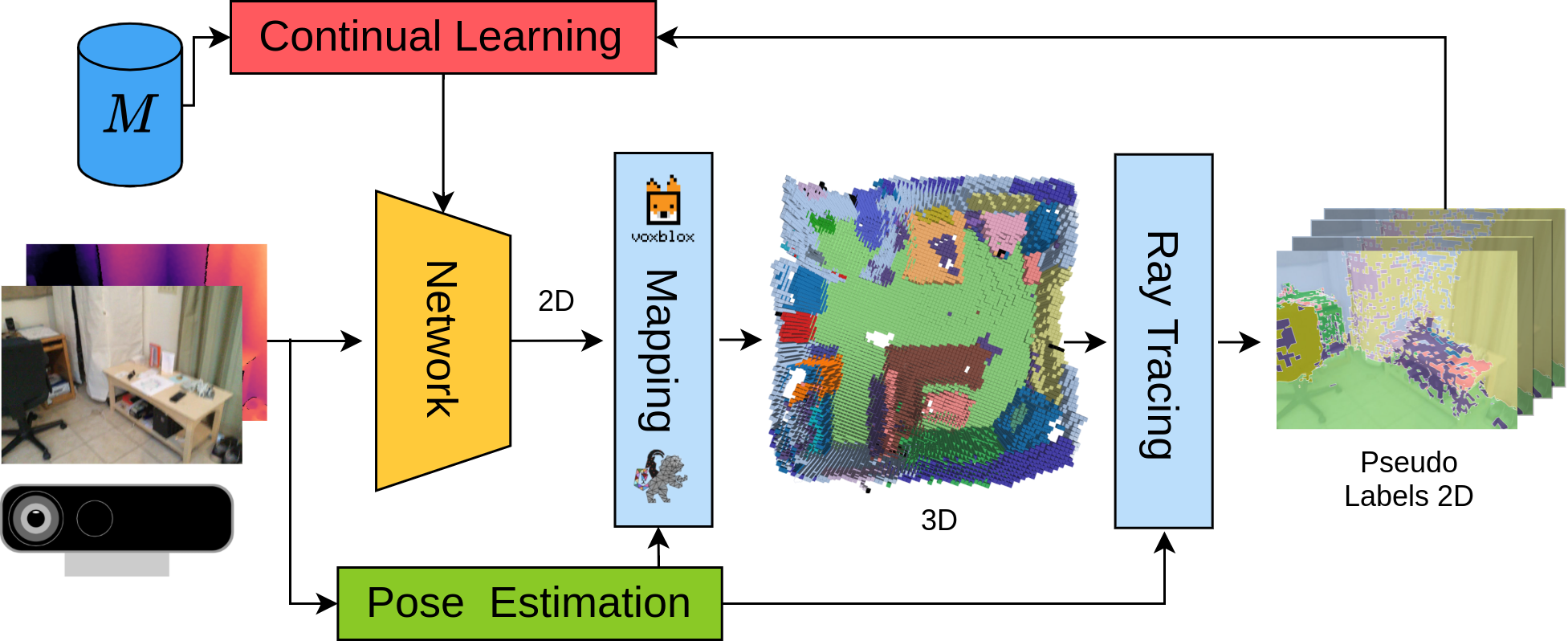}
	\caption{Overview: An RGB-D camera provides \red{inputs} to a segmentation network (yellow) and pose estimation (green). 2D semantic estimates are accumulated in a 3D voxel map. \red{Using ray tracing, we render 2D pseudo-labels from the map}. These are used to adapt the network using a continual learning strategy (red), which can access previously stored samples in a memory buffer (dark blue).}
	\label{fig:draft}
	\vspace{-4mm}
\end{figure}

Our proposed method consists of two main components: 
1.)~The \emph{Pseudo-Label Generation}~(Sec.~\ref{sub:plg}) probabilistically accumulates semantic information in a 3D voxel-based map and generates pseudo-labels using ray tracing. 
2.)~The \emph{Continual Learning}~(Sec.~\ref{sub:cl}) component adapts the parameters of the neural network and receives as input the generated pseudo-labels with the corresponding camera images. 
It is implemented using an experience replay continual-learning strategy and minimizing the cross-entropy between the generated pseudo-labels and the network predictions.
\subsection{Pseudo-Label Generation}
\label{sub:plg}
A pre-trained semantic segmentation network $f_\theta$ predicts initial semantic estimates $Y_n^\textrm{pred}$ from a provided video sequence consisting of individual key frames $I_n$, where $n$ denotes the index in the sequence of length $N$. 
We use Kimera Semantics~\cite{kimera20} to create a dense semantic map of the robot's environment. The geometry of the scene is represented by voxel-based truncated signed distance function~(TSDF). 
In addition to the TSDF volume, a semantic voxel volume stores the probability of each voxel belonging to a semantic class.
A SLAM module~\cite{orbslam16, BundleFusion17} estimates the camera extrinsics $H_n$, which are used together with $D_n$ to integrate the predicted semantics $Y_n^\textrm{pred}$ into both volumes. 
The TSDF is calculated following~\cite{voxblox17}.
For each voxel close to the TSDF surface, the semantic label probability is updated following recursive Bayesian estimation~\cite{kimera20}. 
The predicted semantic labels $Y_n^\textrm{pred}$ are one-hot encoded \red{and used together with the depth image $D_n$ to generate a 3D point cloud, where each point contains the information about its predicted semantic label. 
All the points in the generated cloud that fall into the boundaries of a given voxel are used as new measurements ($Y_n^\textrm{pred}$) to update the stored probability distribution of the voxel belonging to a semantic class.}

This mapping procedure is performed for each camera trajectory within a scene individually.
After integration of all $N$ measurements, Marching Cubes~\cite{marchingcubes87} is used to estimate a high-resolution mesh. 
We ray trace the mesh for each camera pose $H_n$ to determine for each pixel in the camera plane the first intersection of the corresponding ray with the mesh. 
Each of the resulting 3D locations is then used to index the semantic voxel volume in $O(1)$ time to retrieve the semantic label probabilities for the associated pixel. 
We refer to the resulting re-projected semantic segmentation label as \emph{pseudo-label} $Y^\textrm{pseudo}_n$. 
The pseudo-labels aggregate information from multiple viewpoints and enforce multi-view consistency.
At the same time, gathering information from multiple viewpoints allows filtering out semantic segmentation errors induced by bad lighting, motion blur, and outlier predictions in individual frames.
This allows us to generate a learning signal of higher accuracy that can be used to adapt the network in the absence of ground-truth supervision. 
We exploit the learning signal to adapt the pre-trained network's parameters~$\theta$ (Sec.~\ref{sub:cl}).
Exact implementation details are discussed in Section~\ref{sec:implementation}.

\subsection{Continual \red{Adaptation}}
\label{sub:cl}
While it would be possible to directly replace the single-frame prediction of the segmentation network with the pseudo-labels, we instead use the pseudo-labels to train the neural network and adapt it to the scene. This has two reasons. First, the accuracy gain of the pseudo-labels cannot be transferred to a different environment, since the map is bound to the geometry of the scene. The (adapted) network, on the other hand, has the ability to transfer the gained knowledge to any future frame from any environment. Second, the pseudo-labels themselves require sufficient prediction accuracy of the network, while the network training has the potential to filter out undesirable artifacts from the voxel rendering. 

For training, we one-hot encode the pseudo-labels according to the most likely class per pixel.
This experimentally outperformed the probabilistic pseudo-labels and reduces storage and computation needed.
To update the model parameters $\theta$ we use an experience replay strategy.
For this, a small subset of $N_M$ randomly selected samples of pre-training dataset is stored in a memory buffer $M$, which can be accessed to \emph{replay} samples (i.e., feed them again to the network) when adapting the parameters $\theta$ to the current scene. 
The standard cross-entropy loss function is used for both the new samples annotated with the pseudo-labels and the replayed samples with ground-truth annotations.
We use stochastic gradient descent (SGD) to optimize the cross-entropy loss function. 
We can explicitly distinguish between the loss induced by samples stored in the memory buffer and the pseudo-labels in the SGD update:
% equation: 
\begin{align}
l_\textrm{total} &=
\sum_{i=1}^{n_\textrm{pseudo}}\; l_{\textrm{CE}}(f_\theta(x_i),y_i) \;+\; 
	\sum_{j=1}^{n_\textrm{rep}}\; l_{\textrm{CE}}(f_\theta(x_j),y_j)\\
\theta_{t+1} &= \theta_t - \;\frac{\mu}{n_\textrm{pseudo}+n_\textrm{rep}}
		\;\frac{d}{d\theta}  \; l_\textrm{total},
\end{align}

where $n_\textrm{rep}$ and $n_\textrm{pseudo}$ denote the number of replayed and pseudo-labels samples respectively. 
The learning rate is denoted by $\mu$ and $l_{\textrm{CE}}$ is the cross-entropy loss function. 
On one side, minimizing the loss of the replayed samples motivates preservation of previously learned information: the diversity of the samples in the memory buffer favors generalization and mitigates overfitting to the small pseudo-label dataset.
On the other side, the loss of the pseudo-labels encourages the learning of new knowledge.
Additionally, since the samples in the memory buffer are annotated with ground-truth labels, common patterns of the ground-truth annotations may be transferred to the pseudo-labeled samples and act as a regularization mechanism.
Finally, storing a subset of $N_M$ samples significantly reduces the memory needed with respect to the full pre-training dataset size $N_\textrm{pre}$. 
The network training routine is elaborated in Section~{\ref{sec:implementation}}. 

%% file: content/4_Implementation_Details.tex
\section{IMPLEMENTATION DETAILS}
\label{sec:implementation}
\subsection{Network and Dataset}
We use Fast-SCNN~\cite{FastSCNN2019} as a semantic segmentation network.
During inference it runs at over \SI{250}{fps} using a resolution of $320\times640$ pixels with \SI{1.1}{\mega\nothing} parameters. 
For quantitative evaluation, we use the ScanNet~\cite{Scannet17} dataset. 
It consists of 1513 Microsoft Kinect camera trajectories recorded at \SI{30}{fps} within 707 distinct indoor spaces. 
For each scan, the dataset provides a dense 3D map, which is manually annotated with NYU40 classes ~\cite{NYU12} and per-frame labels generated by 2D re-projection. 
The pre-training dataset consists of every $100$th frame of scene 11-707 resulting in \SI{\sim25}{\kilo\nothing} frames. 
From this we use \SI{20}{\kilo\nothing} frames for the actual pre-training and \SI{5}{\kilo\nothing} for testing the performance on the pre-training dataset. 
All scans recorded in scenes 1-5 are used to evaluate the adaptation performance of our proposed method. 
For each scene, up to 3 separate video sequences are provided in the dataset. 
\red{From each sequence}, the first $80\%$ of the frames are used for pseudo-label generation and continually training the network. 
The final $20\%$ of the frames are only used for testing the adaptation performance on novel views of the same scene.
%Despite the strict training-test split, the same objects may be observed within the same scene in both datasets. 
We stress that the created benchmark mimics a real robotic scenario, in which a large dataset of annotated data is commonly available, but adaptation during a mission has to be performed in an unsupervised manner. 

\subsection{Network Pre-Training}
For pre-training, we use Adam~\cite{Adam14} with a batch size of 8. 
The learning rate starts at $10^{-3}$ and polynomially decays over 150 epochs to $10^{-6}$ with a rate of 0.9. 
We stop the training procedure early after 65 epochs ($\sim200.000$ optimization steps) given convergence on a test set. 
During training, we apply standard data augmentation, including color jitter, horizontal flipping, and random cropping.

\subsection{Pseudo-Label Generation}
To construct the semantic map, we use the provided implementation by Kimera Semantics~\cite{kimera20}, which builds on Voxblox~\cite{voxblox17}, a mapping framework based on voxel grids. 
We set the voxel resolution to \SI{3}{\centi\meter}, which we found to provide a good balance between level of semantic detail captured and computational efficiency.
Kimera Semantics tracks the full posterior probability for all 40 NYU40 labels per voxel.
Integration of a single measurement into the TSDF and semantic volume on a Ryzen 5900X CPU takes \SI{330}{ms} at a resolution of $320\times640$. 
For typical room-sized scenes \SI{10}-\SI{20}{\meter\squared} Marching Cubes produces a mesh of around \SI{5}{MB}. 
The voxel volume storing the full semantic posterior has a size of $\sim$\SI{500}{MB} for 4.1x3.6x1.5 m volume~(Fig.~\ref{fig:scannet_maps}). 
The high-performance CPU-based ray tracing implementation infers pseudo-labels at a rate of \SI{30}{fps}.

\subsection{Continual Learning}
\label{subsec:imp_cl}
We retrain the network on the generated pseudo-labels for a total of 50 epochs. 
SGD is used with a 1cycle learning rate schedule~\cite{OneCycleLR17} and a batch size of $8$ to adapt the parameters. 
The scheduler linearly increases the learning rate from $10^{-6}$ to $0.05$ over the initial 5 epochs and successively decays it to $10^{-3}$ over the remaining 45 epochs. 
Starting with a slow learning rate is important to avoid strongly perturbing the model parameters within the first iterations of training. 
We empirically found that storing $10$\% of the pre-training dataset in the memory buffer (resulting in a memory size $N_M$ of $2000$) is capable of representing the training dataset adequately. 
During training the samples of each mini-batch are randomly chosen with a ratio of 4:1 from the pseudo-labels and memory buffer. 
We experimentally found this ratio to provide a good trade-off between integrating new knowledge and preserving the performance on the pre-training dataset. 
During training the same data augmentation used for pre-training is applied to the replayed and pseudo-labeled samples. 
We found data augmentation to be particularly beneficial for small buffer sizes, which aligns with the findings reported in~\cite{rethinking_er20}. 

%% file: content/5_Experiments.tex
\section{EXPERIMENTS}
In the following, we evaluate each component in the pipeline individually to measure its performance. 
We then test the fully operating pipeline and show experimental results for the ScanNet dataset and data recorded in a real-world indoor scene using a handheld RGB-D sensor. 

\label{sec:experiments}
\subsection{Pseudo-Label Performance}
\label{subsec:pl}
\input{content/figures/scannet_matrix}
To evaluate the pseudo-label generation procedure, we compare the segmentation accuracy achieved by the pre-trained neural network to that of the resulting pseudo-labels.
We illustrate 4 frames of the first scene in Figure~\ref{fig:scannet_matrix}. 
The pre-trained network predictions disagree for the same location over consecutive frames. 
The pseudo-labels\red{, on the other hand,} include minor artifacts induced by the voxel discretization and the ray tracing process, but are consistent over multiple iterations. 
Moreover, as we show in Section~\ref{subsec:cl}, these artifacts are not reflected in the adapted network predictions and do not prevent the signal from being beneficial for improving the prediction accuracy.
To verify the correctness of the pseudo-label generation and set an upper bound for the pseudo-label performance, we additionally use the ground-truth segmentation to generate pseudo-labels. 

\red{As metrics for our evaluations we use primarily the total accuracy (Acc), and additionally report mean Intersection of Union (mIoU).}
\red{We report each metric on the test dataset both per} scene and averaged over all scenes.
\begin{table*}[t]

	\centering
	\setlength{\tabcolsep}{3pt}
	\ra{1.2}
	\footnotesize
	\begin{tabular}{cllllcrcrcrcrcrcrcrcrclccccc}\toprule
     &      &     \multicolumn{2}{c}{Frames}     && \multicolumn{16}{c}{Adaptation} &&  \multicolumn{6}{c}{Generalization} \\
         \cmidrule{3-4}      \cmidrule{6-22}  \cmidrule{24-28}
& Scene   & Train & Test && 1-Pred && 1-Pse  && URMA && 2-FT && 2-Pred && 2-Pse && 3-Pred && 3-Pse && GT-Pse && 1-Pred & URMA & 2-FT & 2-Pred & 3-Pred \\ \midrule
\multirow{6}{*}{\rotatebox{90}{accuracy}} &
1             &	1415 & 352 && 61.7   && 66.1   && 62.8 && 66.9 && 66.9 && \underline{68.6} && \textbf{67.1} && 68.0 && 94.8  && 60.7 & 54.5  & 47.3  & \textbf{59.5} & 58.3 \\
& 2             & 227  & 57  && 52.4   && 50.1   && 53.9 && 55.8 && \textbf{59.4} && \underline{52.4} && 52.1 && 51.2 && 94.8  && 60.7 & 51.9  & 44.9  & \textbf{59.4 }& 58.5 \\
& 3             &	997  & 249 && 38.3   && 40.5   && 37.0 && 39.8 && \textbf{41.1} && \underline{41.9} && 41.0 && 40.8 && 93.6  && 60.7 & 55.6  & 42.4  & \textbf{59.9} & 58.0 \\
& 4             &	381  & 96  && 66.1   && \underline{78.1}   && 62.9 && 74.9 && \textbf{75.2} && 76.5 && 74.8 && 75.9 && 97.2  && 60.7 & 51.5  & 40.2  & \textbf{58.6} & 58.1 \\
& 5             &	75   & 18  && 38.8   && 39.6   && 36.6 && 40.2 && \textbf{40.4} && 49.0 && 39.6 && \underline{49.6} && 84.5  && 60.7 & 40.9  & 36.9  & \textbf{53.5} & 48.4 \\  \cmidrule{2-28}
& \textbf{AVG}  &-     &-    && 51.5   && 54.9   && 50.6 && 55.5 && \textbf{56.6} && \underline{57.7} && 54.9 && 57.1 && 93.0  && 60.7 & 50.9  & 42.3  & \textbf{58.2} & 56.3 \\   \midrule 
\multirow{6}{*}{\rotatebox{90}{mIoU}} & 1             &	1415 & 352 && 21.9   && 19.8  && \textbf{24.6} && 21.4 && 21.5 && 22.9 && 22.1 && 22.6 && 81.5  && 28.2 & 24.1  & 11.0  & \textbf{27.6} & 26.8 \\
& 2             & 227  & 57  && 23.0   && 19.8  && \textbf{23.8} && 21.2 && 21.4 && \underline{20.4} && 19.8 && 19.9 && 77.1  && 28.2 & 21.3  & 10.6  & \textbf{27.5} & 26.7 \\
& 3             &	997  & 249 && 13.2   && 13.6  && 12.0 && 11.9 && \textbf{13.5} && \underline{13.7} && 12.5 && 13.6 && 83.3  && 28.2 & 24.0  & 7.9   & \textbf{27.4} & 25.8 \\
& 4             &	381  & 96  && 33.7   && 49.5  && 30.1 && 42.7 && \textbf{45.7} && \underline{48.3} && 45.6 && 47.6 && 90.9  && 28.2 & 30.9  & 9.3   & \textbf{26.2} & 26.3 \\
& 5             &	75   & 18  && 24.1   && 27.9  && 28.3 && \textbf{31.0} && 30.8 && 34.4 && 29.0 && \underline{34.9} && 67.2  && 28.2 & 17.4  & 7.4   & \textbf{23.0} & 20.0 \\ \cmidrule{2-28}
& \textbf{AVG}  &-     &-    && 23.2   && 26.1  && 23.8 && 25.6 && \textbf{26.6} && \underline{27.9} && 25.8 && 27.7 && 80.0  && 28.2 & 23.5  & 9.2   & \textbf{26.3} & 25.1 \\   \bottomrule

	\end{tabular}
		\caption{\red{Segmentation results: Top rows (Acc), bottom rows (mIoU); Methods are evaluated on the adaptation to novel scenes~1-5 of the ScanNet test dataset and Generalization performance on the pre-training test dataset; Methods: pre-trained network~(1-Pred), URMA-baseline (URMA~\cite{2021UncertaintyReduction}), fine-tuned network (2-FT), continually-learned networks (2-Pred and 3-Pred), pseudo-labels generated from the pre-trained network (1-Pse), or from the continually-learned networks (2-Pse and 3-Pse), and pseudo labels based on ground truth as an upper bound (GT-Pse). In bold the best performing network is indicated and the best generated pseudo label is underlined.}}
	\label{tab:verify}
\vspace{-0.3cm}
\end{table*}
As shown in Table~\ref{tab:verify}, the pseudo-labels (\mbox{1-Pse}) generated based on the pre-trained network predictions (\mbox{1-Pred}) improve the accuracy, on average \red{from 51.5\% to 54.9\%, a relative increase of 6.6\%}. 
\red{While for scene 2 and 5 the pseudo-labels accuracy is lower (-2.3\% in scene 2) or not significantly higher (+0.8\% in scene 5) compared to that of the pre-trained (1-Pred) model, on the training dataset the accuracy is 60.6\% and 64.1\% for scene 2 and scene 5 respectively, making it a viable training signal. }
Since the artifacts induced by voxelization and ray tracing limit the accuracy of the pseudo-labels, we also report as an upper bound the accuracy of the pseudo-labels generated by integrating into the map the ground-truth data (\mbox{GT}) instead of the network predictions. 
We show the reconstructed mesh used to generate \mbox{1-Pse} and \mbox{GT-Pse} for an example scene in Figure~\ref{fig:scannet_maps}. 
As visible from the right image, all the objects can be clearly identified in the GT~map. 
\red{Despite a small number of misclassifications in \mbox{1-Pse} (\emph{desk}, top right; \emph{toilet} top left) and artifacts (\emph{sofa} bottom middle, \emph{bed} top middle) that cannot be resolved, the overall scene is segmented correctly. 
Furthermore, the quantitative results reported in the next Sections demonstrate that enforcing view-consistency using a 3D representation and reprojecting the aggregated information to 2D provides a suitable learning signal for network adaptation.}

\red{We observed that our pseudo label generation method favours the reconstruction of bigger objects or classes within the scene. This may be attributed to the fact that small objects are commonly observed from fewer viewpoints, thus providing less data to enforce multi-view consistency.}

\input{content/figures/scannet_maps}

\subsection{Continual Learning}
\label{subsec:cl}
We now analyse how the generated pseudo-labels can be used to integrate the gained domain-specific knowledge into the segmentation network. 
Specifically, we are interested in how much the segmentation network can improve its performance on the current scene, and how much it loses generalisation to other scenes, essentially posing the adaptation task as a continual learning problem. 
In a simpler binary segmentation setting, \cite{Blum21} evaluated a range of continual learning methods, and found that the strongest competitor to experience replay is na\"ive fine-tuning, which we therefore also compare here.

\red{In addition to na\"ive fine-tuning we compare our approach to URMA~\cite{2021UncertaintyReduction}, an UDA method for semantic segmentation. 
We highlight that UDA methods in general do not tackle the problem of integrating additional knowledge into the network, but focus instead on transferring knowledge from one domain to the other. Additionally, they are mostly studied in an offline scenario.
We adapt the original authors' implementation by replacing the segmentation network with FastSCNN~\cite{FastSCNN2019} for a fair comparison with our method. 
The authors choose a different number of training epochs depending on the dataset. 
We found that the performance of URMA strongly varies across training epochs and we could not determine a single training epoch that leads to good results across scenes. 
For this reason, we provide an additional benefit to URMA~\cite{2021UncertaintyReduction} by reporting the best performing epoch during 10 training epochs.}
In general, providing a fair benchmark that takes into account all the design parameters (e.g., number of update steps, number of samples, memory, time, compute) is extremely challenging in the field of continual learning, making results often highly dependent on the chosen method or benchmark~\cite{Lesort2020CLForRobotics}.
The na\"ive fine-tuning is implemented using the same learning procedure as for the continual learning approach, described in Section~\ref{subsec:imp_cl}, but without replaying any samples from the memory buffer. 
During continual learning and fine-tuning the same total number of new samples are provided to the network. 

We evaluate the broader generalization performance \red{(Gen)} by measuring the accuracy on the test split of the pre-training dataset and we calculate the adaptation performance using the \red{test dataset of the respectively adapted scene for} scene 1-5 of the ScanNet dataset.
As shown in Table~\ref{tab:verify}, the network adapted using our method (2-Pred) outperforms the pre-trained network in terms of adaptation (\red{Acc $+5.1$\%}). 
Moreover, the average accuracy \red{of our method} slightly increases over na\"ive fine-tuning \red{(by $+1.1$\%, from 2-FT 55.5\% to 2-Pred 56.6\% in Acc)}.
We hypothesize that for some scenes the samples stored within the memory buffer induce a positive forward transfer \red{by regularizing the adaptation procedure. For instance, by providing the ground labels for the replayed samples the network may transfer properties from the ground-truth labels (e.g., their smoothness) to the current scene, where only pseudo labels with artifacts are available.} 
However, given that this improvement is not significant, we conclude that continual learning does not hinder adaptation compared to fine-tuning. 
\red{In addition, the continually-trained network (2-Pred) outperforms the generated pseudo labels in adaptation performance. To understand this effect, consider that we measure performance on the validation split of the scene, whereas 2-Pred is trained on the pseudo labels from the training split, potentially having a higher accuracy given the longer sequence (Sec.~\ref{subsec:pl}). This result shows that the adapted network can transfer knowledge gained from the training dataset to the test dataset, where semantic mapping and reprojection cannot improve the performance.%
}
The predicted semantic segmentation of the continually-learned network for the selected key-frames of the first scene is illustrated in Figure~\ref{fig:scannet_matrix}. 
As evident in the third row, our methods predictions align with the pseudo-labels but filter out noise and artifacts, resulting in smooth boundary regions.

As expected, the continual learning strategy (\red{Gen. Acc $58.2\%$}) is capable of mitigating catastrophic forgetting compared to fine-tuning (\red{Gen. Acc $42.3\%$) and URMA (Gen. Acc $50.9\%$)}, but \red{minor} forgetting still occurs compared to the pre-trained network (\red{Gen. Acc $60.7\%$}). 
Our evaluation does not reveal a correlation between the number of frames in a scene and the achieved adaptation performance. 
Overall, the mIoU metric varies more across scenes than the accuracy metric, which we suppose is related to the pixel-wise class imbalance within each scene, to the fact that we use the unbalanced cross-entropy loss function, as well as to the possibly weaker supervision signal for small objects~(Sec.~\ref{subsec:pl}). 
Therefore, while the averaged mIoU metric over all scenes increases for our method, the mIoU decreases slightly for 2 out of 5 scenes. 
In general the mIoU is sensitive to classes that are only presented within a few frames.

UMRA~\cite{2021UncertaintyReduction} outperforms in 2 out of 5 scenes the pre-trained network, but overall performs worse (Adap Acc 50.6\%) than the pre-trained network (Adap Acc 51.5\%) and significantly worse compared to our continually-learned network (Adap Acc 56.6\%) in accuracy. 
In terms of mIoU, URMA~\cite{2021UncertaintyReduction} outperforms our method on scene 1 and 2, when provided with the added benefit of selecting from 10 possible models per scene. 
We believe that the overall performance increase of our method with respect to~\cite{2021UncertaintyReduction} can be traced back to the generation process of the supervision signal used to update the network and to the availability of a small replay dataset. 
Our method exploits the fact that in a robotic scenario information is captured from different viewpoints and we therefore create a supervision signal based on the fused network predictions. On the other hand,~\cite{2021UncertaintyReduction} creates a supervision signal on a per-image basis, which makes it applicable to non-sequential data, but does not exploit the constraint of temporal or multi-view consistency.

\input{content/figures/scannet_certainty}
We can evaluate \red{this} multi-view consistency of the network predictions before and after adaptation.
When integrating disagreeing network predictions in the same voxel, the uncertainty of the specific voxel increases.
Figure~\ref{fig:scannet_certainty} shows these voxel uncertainties after integrating predictions from the pre-training network (\mbox{1-Pred}) and the adapted continually-learned network (\mbox{2-Pred}). 
\red{We note that the presented uncertainty only depends on the one-hot encoded label predictions of the network and no information about uncertainty for a single-image prediction can be inferred.}
Clearly, our adaptation procedure increases the overall certainty and therefore multi-view consistency.

\subsection{Iterative Operation}
The process of generating pseudo-labels and adapting the neural network using continual learning can also be performed for multiple steps within the same scene, by iteratively generating pseudo-labels and retraining the network on these.
We hypothesized that iterative adaptation by consecutively transforming between data representations from 2D to 3D could increase the performance further given that the labels used to generate the 3D map are more accurate. 
We report the \red{results similarly to Section~\ref{subsec:cl} in Table~\ref{tab:verify}}. The pseudo-label generation \red{(2-Pse, 3-Pse)} and network training \red{(3-Pred)} is performed strictly following the procedure elaborated in Section~\ref{sec:implementation} for all iterations.

As shown in Table~\ref{tab:verify}, after the first adaptation step the network accuracy does not improve for four of the five scenes tested. 
We reason that after the first iteration the adapted network already aligns with the multi-view consistency constraint. 
Therefore, remapping the multi-view consistent labels into 3D cannot resolve disagreeing semantic estimations and potentially introduces discretization artifacts.
This leads us to the conclusion that a single iteration of mapping and continual learning is the most effective for achieving a positive network adaptation \red{while mitigating forgetting}.

\subsection{Deployment on Handheld Device}
To test our proposed method in the wild, we capture data of multiple scenes with a hand-held Azure Kinect \mbox{RGB-D} sensor in different office spaces. 
The network pre-trained on ScanNet is used to estimate an initial semantic segmentation of the captured data.
We use the open-source RGB-D SLAM system ORB-SLAM2~\cite{orbslam16} to retrieve the camera poses after bundle adjustment and loop closures. We then build the volumetric map using these poses.
The Azure Kinect sensor cannot measure depth for reflective and light-absorbing surfaces and is limited to a maximum distance of \SI{5.45}{m}. 
Figure~\ref{fig:labdata_matrix} shows examples of the resulting pre-training, pseudo-label and adapted network predictions. 
As clear from the top row, the pre-trained network misclassifies multiple objects (\emph{table}, \emph{floor}). 
This evidence is in line with the significant distribution mismatch between the recorded data and the pre-training dataset, which was recorded with a different sensor. 
The generated pseudo-labels (1-Pse) correctly classify the \emph{desk} in all frames. 
Given the light-absorbing carpet and reflective television, no depth measurements can be integrated into the volumetric map, leading to a semi-dense mapping, which induces undefined semantics when ray tracing the pseudo-labels. 
When training the network, we default to the \mbox{1-Pred} predictions for these pixels with undefined semantics in the pseudo-labels \mbox{1-Pse}. 
This allows effectively avoiding training on a sparse supervision signal, which would lead the network to wrongly classify not mapped regions (\emph{floor}, \emph{television}) with the label of the closest mapped pixels.
We conclude that our method generalizes well to this less-controlled deployment scenario, showing the suitability of our approach for real-world robotic applications.

%% file: content/figures/scannet_matrix.tex
\begin{figure}[t]
	\caption*{\footnotesize{1-Pred: Prediction of Pre-trained Network}}
	\begin{minipage}[t]{0.24\linewidth}
		\vspace{-0.25cm}
		\includegraphics[width=\textwidth]{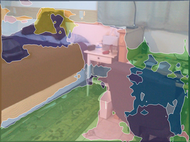}
	\end{minipage}
	\hfill
	\begin{minipage}[t]{0.24\linewidth}
		\vspace{-0.25cm}
		\includegraphics[width=\textwidth]{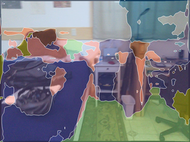}
	\end{minipage}
	\hfill
	\begin{minipage}[t]{0.24\linewidth}
		\vspace{-0.25cm}
		\includegraphics[width=\textwidth]{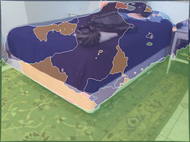}
	\end{minipage}
	\hfill
	\begin{minipage}[t]{0.24\linewidth}
		\vspace{-0.25cm}
		\includegraphics[width=\textwidth]{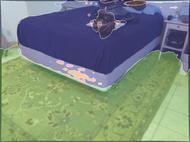}
	\end{minipage}
	
	\vspace{-0.05cm}
	
	\caption*{\footnotesize{1-Pse: Pseudo-Labels from Semantic Map}}
	\begin{minipage}[t]{0.24\linewidth}
		\vspace{-0.25cm}
		\includegraphics[width=\textwidth]{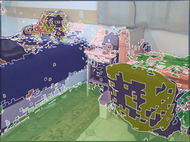}
	\end{minipage}
	\hfill
	\begin{minipage}[t]{0.24\linewidth}
		\vspace{-0.25cm}
		\includegraphics[width=\textwidth]{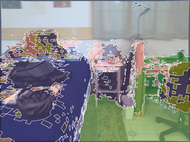}
	\end{minipage}
	\hfill
	\begin{minipage}[t]{0.24\linewidth}
		\vspace{-0.25cm}
		\includegraphics[width=\textwidth]{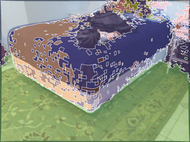}
	\end{minipage}
	\hfill
	\begin{minipage}[t]{0.24\linewidth}
		\vspace{-0.25cm}
		\includegraphics[width=\textwidth]{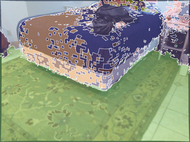}
	\end{minipage}
	
	\vspace{-0.05cm}
	\caption*{\footnotesize{2-Pred: Prediction of Adapted Network Trained with 1-Pse}}	
	\begin{minipage}[t]{0.24\linewidth}
		\vspace{-0.25cm}
		\includegraphics[width=\textwidth]{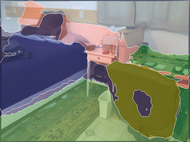}
	\end{minipage}
	\hfill
	\begin{minipage}[t]{0.24\linewidth}
		\vspace{-0.25cm}
		\includegraphics[width=\textwidth]{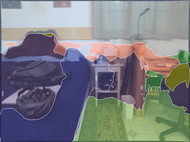}
	\end{minipage}
	\hfill
	\begin{minipage}[t]{0.24\linewidth}
		\vspace{-0.25cm}
		\includegraphics[width=\textwidth]{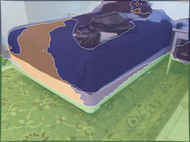}
	\end{minipage}
	\hfill
	\begin{minipage}[t]{0.24\linewidth}
		\vspace{-0.25cm}
		\includegraphics[width=\textwidth]{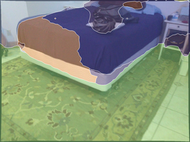}
	\end{minipage}
	
	\vspace{-0.05cm}
	\caption*{\footnotesize{Ground Truth}}
    \begin{minipage}[t]{0.24\linewidth}
		\vspace{-0.25cm}
		\includegraphics[width=\textwidth]{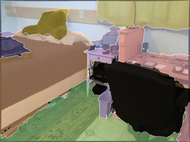}
	\end{minipage}
	\hfill
	\begin{minipage}[t]{0.24\linewidth}
		\vspace{-0.25cm}
		\includegraphics[width=\textwidth]{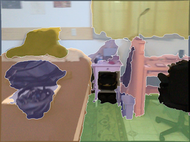}
	\end{minipage}
	\hfill
	\begin{minipage}[t]{0.24\linewidth}
		\vspace{-0.25cm}
		\includegraphics[width=\textwidth]{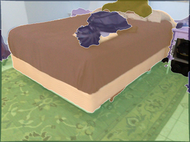}
	\end{minipage}
	\hfill
	\begin{minipage}[t]{0.24\linewidth}
		\vspace{-0.25cm}
		\includegraphics[width=\textwidth]{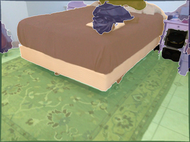}
	\end{minipage}
	%\vspace{1mm}
	\caption{Segmentation of the first scene in the ScanNet dataset, using ScanNet color coding. First row: Pre-trained network predictions (1-Pred). Second row: Generated pseudo-labels (1-Pse). Third row: Adapted neural network (2-Pred). Fourth row: Ground-truth labels.}
	\label{fig:scannet_matrix}
	\vspace{-0.6cm}
\end{figure}

%% file: content/figures/scannet_maps.tex
\begin{figure}[t]
	\begin{minipage}[t]{0.49\linewidth}
		\caption*{\footnotesize{1-Pse}}
		\vspace{-0.25cm}
		\includegraphics[width=\textwidth]{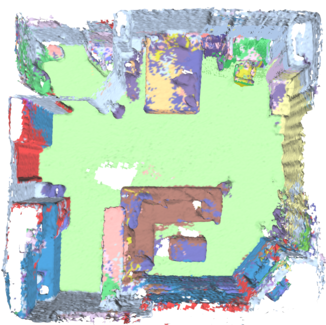}
	\end{minipage}
	\hfill
	\begin{minipage}[t]{0.49\linewidth}
		\caption*{\footnotesize{GT-Pse}}
		\vspace{-0.25cm}
		\includegraphics[width=\textwidth]{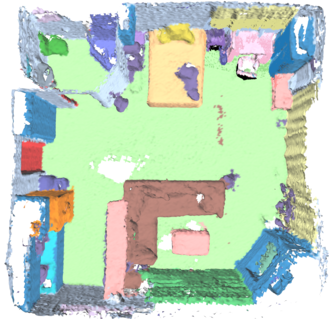}
	\end{minipage}
	\caption{Meshes used for pseudo-label generation of the first ScanNet scene. Left: Pseudo-label map (1-Pse) generated using the pre-trained neural network. Right: Pseudo-label map (GT-Pse) generated using the ground-truth labels.}
	\label{fig:scannet_maps}
	\vspace{-5mm}
\end{figure}

%% file: content/figures/scannet_certainty.tex
\begin{figure}[t]
	\begin{minipage}[t]{0.47\linewidth}
		\caption*{\footnotesize{Consistency of 1-Pred}}
		\vspace{-0.25cm}
		\includegraphics[width=\textwidth]{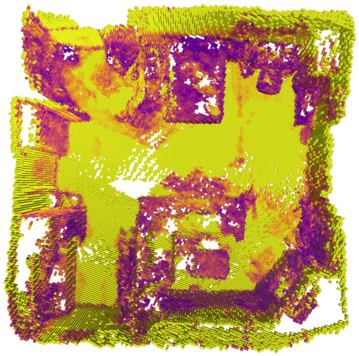}
	\end{minipage}
	\hspace{-0.2cm}
	\begin{minipage}[t]{0.05\linewidth}
	    \vspace{1cm}
		\includegraphics[width=\textwidth]{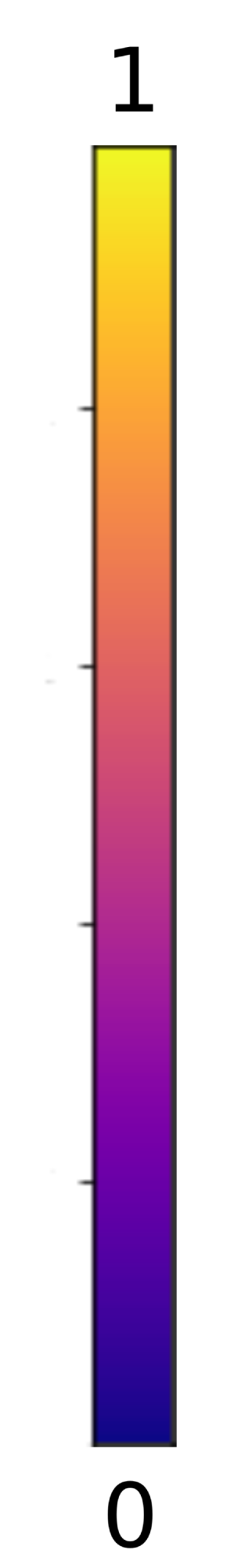}
	\end{minipage}
	\hspace{-0.2cm}
	\begin{minipage}[t]{0.47\linewidth}
		\caption*{\footnotesize{Consistency of 2-Pred}}
		\vspace{-0.25cm}
		\includegraphics[width=\textwidth]{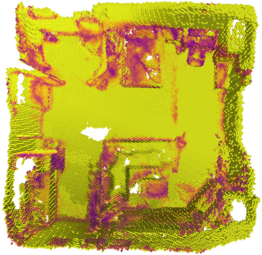}
	\end{minipage}
	\caption{Multi-view consistency measured as per-voxel confidence in the first ScanNet scene. Left: Confidence of the voxel volume when mapping the pre-trained neural network predictions (1-Pred). Right: Confidence of the voxel volume when mapping the adapted neural network predictions in the second iteration (2-Pred).}
	\label{fig:scannet_certainty}
	\vspace{-0.7cm}
\end{figure}

%% file: content/6_Conclusion.tex
\section{CONCLUSION}
We showed that leveraging complementary 2D-3D data representations creates a useful learning signal for semantic segmentation without any external supervision. 
To the best of our knowledge, we are the first to apply a continual learning strategy to adapt a multi-class semantic segmentation network in a robotic mission scenario. 
Our experiments show that a 3D data representation that intrinsically enforces multi-view consistency can be effectively used to retrain a network to comply with this consistency constraint already within one iteration. 
When evaluated on a real-world indoor dataset, our method increases the semantic segmentation performance on average by \red{9.9}\% relative to the pre-trained network. Further experiments with hand-held sensor recordings show how our proposed approach can be applied in a real-world scenario. In conclusion, we show a ready-to-deploy continual learning approach for semantic segmentation that does not require any prior knowledge of the scene or any external supervision and can simultaneously retain knowledge of previously seen environments while integrating new knowledge.

Our approach requires volumetric mapping, which is so far limited by the assumption of a static world and by the operating characteristics of the existing RGB-D sensors. 
Furthermore, consistent misclassifications within a scene cannot be resolved by multi-view consistency. 
Future work may extend our approach by including prior knowledge, e.g., identification of object instances or application of scene understanding methods to potentially resolve more misclassifications. 
Investigation of the effects of incorrect feedback loops caused by a wrong self-supervision signal can provide insights on how to prevent consistent misclassifications. 
\red{Additionally, a promising future direction may be to investigate using the prior of multi-view consistency enforced by the global 3D map as a self-supervision method for representation learning. }
Based on our work, further research can be conducted to reduce the gap between fundamental continual learning research and real-world robotic applications.